%% file: main.tex
\PassOptionsToPackage{table}{xcolor}
\documentclass[10pt,twocolumn,letterpaper]{article}

\usepackage{tikz}
\usepackage{circuitikz}
\usepackage{makecell}
\usepackage{multirow}
\usepackage{afterpage}
\usepackage{colortbl}

\input{sections/macros}

\usepackage[]{cvpr}

\definecolor{cvprblue}{rgb}{0.21,0.49,0.74}
\usepackage[pagebackref,breaklinks,colorlinks,citecolor=cvprblue]{hyperref}

\def\paperID{410} %
\def\confName{3DV\xspace}
\def\confYear{2026\xspace}

\hypersetup{
    colorlinks=true,
    linkcolor=darkblue,
    citecolor=darkblue,
    filecolor=darkblue,
    urlcolor=darkblue
}

\definecolor{darkblue}{RGB}{0,51,102}

\definecolor{somegray}{rgb}{0.5, 0.5, 0.5}
\newcommand{\darkgrayed}[1]{\textcolor{somegray}{#1}}
\makeatletter
\newcommand*\titleheader[1]{\gdef\@titleheader{#1}}
\AtBeginDocument{%
  \let\st@red@title\@title
  \def\@title{%
    \vskip-3.5em
    \bgroup\normalfont\large\centering\@titleheader\par\egroup
    \vskip1.5em\st@red@title}
}

\makeatother

\titleheader{\darkgrayed{This paper has been accepted for publication at the IEEE International Conference on 3D Vision (3DV), Vancouver, CA, 2026. \copyright IEEE}}

\title{VibES: Induced Vibration for Persistent Event-Based Sensing}

\author{
\begin{tabular}{c@{\hspace{2em}}c@{\hspace{2em}}c}
Vincenzo Polizzi$^{1}$ & Stephen Yang$^{1,2}$ & Quentin Clark$^{3}$ \\
Jonathan Kelly$^{1}$ & Igor Gilitschenski$^{3}$ & David B.\ Lindell$^{3}$
\end{tabular}\\
\\
$^{1}$University of Toronto, Robotics Institute \\
$^{2}$University of Toronto, Department of Mechanical and Industrial Engineering \\
$^{3}$University of Toronto, Department of Computer Science \\
{$^{1}$\tt\small \{vincenzo.polizzi,jonathan.kelly\}@robotics.utias.utoronto.ca} \\
{$^{2}$\tt\small styang@mie.utoronto.ca},
{$^{3}$\tt\small \{qtcc,gilitschenski,lindell\}@cs.toronto.edu}
}

\begin{document}
\maketitle
\input{sections/00_abstract}

\noindent\textbf{Supplementary Material:} For code and data please visit \url{https://papers.starslab.ca/vibes/}.

\input{sections/01_introduction}

\input{sections/02_related}
\input{sections/03_method}

\input{sections/04_experiments}

\input{sections/05_applications}

\input{sections/06_conclusion}

\section*{Acknowledgments}
This work was supported by the Natural Sciences and Engineering Research Council of Canada (NSERC) through the RGPIN program and the Canada Research Chairs program, the Canada Foundation for Innovation, and the Ontario Research Fund.

\clearpage
\appendix
\input{sections/07_supplementary}

{
    \small
    \bibliographystyle{ieeenat_fullname}
    \bibliography{main}
}

\end{document}

%% file: sections/macros.tex
\newcommand{\wframe}{\ensuremath{\mathcal{W}}}
\newcommand{\pointw}{\ensuremath{\mathbf{P}_W}}
\newcommand{\cam}{\ensuremath{\mathcal{C}'}}
\newcommand{\camvir}{\ensuremath{\mathcal{C}}}

\newcommand{\pointcplane}{\ensuremath{\mathbf{p}'}}
\newcommand{\pointcplanevir}{\ensuremath{\mathbf{p}}}

\newcommand{\omegahat}{\ensuremath{\hat{\omega}}}
\newcommand{\tcvirw}{\ensuremath{\mathbf{T}_{CW}}}

\newcommand{\tccvir}{\ensuremath{\mathbf{T}(t)_{C'C}}}

\newcommand{\methodname}[0]{VibES\xspace}
\newcommand{\methodnameshort}[0]{VibES\xspace}

\newcommand{\prophesee}[0]{Prophesee\xspace}

\definecolor{rowgray}{gray}{0.95} 
\definecolor{rowblue}{HTML}{9FC0F2} %

\newcommand{\colortab}[0]{\cellcolor{rowblue!40}}
\newcommand{\colorrow}{\rowcolor{rowblue!40}}

%% file: sections/00_abstract.tex
\begin{abstract}
Event cameras are a bio-inspired class of sensors that asynchronously measure per-pixel intensity changes. Under fixed illumination conditions in static or low-motion scenes, rigidly mounted event cameras are unable to generate any events and become unsuitable for most computer vision tasks.
To address this limitation, recent work has investigated motion-induced event stimulation, which often requires complex hardware or additional optical components. 
In contrast, we introduce a lightweight approach to sustain persistent %
event generation by employing a simple rotating unbalanced mass to induce periodic vibrational motion. This is combined with a motion-compensation pipeline that removes the injected motion and yields clean, motion-corrected events for downstream perception tasks.
We develop a hardware prototype to demonstrate our approach and evaluate it on real-world datasets. 
Our method reliably recovers motion parameters and improves both image reconstruction and edge detection compared to event-based sensing without motion induction.
\end{abstract}

%% file: sections/01_introduction.tex
\section{Introduction}\label{sec:introduction}
\input{figs/qualitative_eval_method}
Most computer vision algorithms rely on conventional image sensors---such as RGB or greyscale cameras---that capture synchronous image frames at fixed intervals. While widely adopted, these frame-based sensors suffer from well-known limitations, including motion blur, latency due to frame timing, and high power consumption~\cite{gallego2022tpami}. These drawbacks are especially problematic in robotics applications where fast dynamics, low latency, and energy efficiency are critical, for example in navigation~\cite{shariff2024event}, feature tracking~\cite{messikommer2023data}, and real-time scene understanding~\cite{kong2024openess}.

Event cameras have emerged as a promising alternative. These bio-inspired sensors asynchronously detect per-pixel changes in log-intensity, producing a sparse stream of events with microsecond latency and high dynamic range~\cite{gallego2022tpami}. Event cameras inherently reduce motion blur, operate at lower power~\cite{gehrig2024low}, and enable low-latency perception, making them well-suited for resource-constrained, high-speed robots. 

However, event cameras do have a fundamental limitation: event generation depends on motion. In static scenes, or when edges align with the direction of the camera motion, events are not triggered. This results in event sparsity, loss of spatial information, and perceptual fading~\cite{clarke1960study}, which undermines long-term feature tracking, edge detection, and high-quality image reconstruction~\cite{botao2024microsaccade}. Consequently, the performance of event-based systems degrades significantly without sufficient camera or scene motion.

Interestingly, the human visual system faces a similar challenge. To prevent image fading during fixation, our eyes perform rapid, involuntary movements known as microsaccades or fixational eye movements (FEMs). These movements continuously stimulate photoreceptors, refreshing the visual input and maintaining high-resolution perception~\cite{krekelberg2011microsaccades, wu2023fixational}. This biological mechanism suggests that artificial motion could be exploited to sustain event generation.

In this work, we draw inspiration from microsaccades while adopting a deterministic, engineered approach. Instead of attempting to replicate their stochastic nature, we design a lightweight vibration mechanism that mechanically stimulates the event camera, ensuring continuous event generation in static or quasi-static scenes. Our approach, \methodname, uses a simple rotating unbalanced mass within a spring–damper system to induce harmonic motion. A motion-compensation pipeline then removes the injected vibration, producing clean, motion-corrected events for downstream tasks. Compared to existing microsaccade-inspired methods based on optical systems~\cite{botao2024microsaccade} or pan–tilt actuators~\cite{d2025wandering,testa2023active,yousefzadeh2018pantilt,Orchard15fns,testa2020dynamic}, our design requires no additional optical components or position-encoding sensors, enables real-time motion compensation on the event stream, and extends to tracking motion frequencies in a target scene. An overview of the proposed method is shown in~\autoref{fig:qualitative_eval_method}. %

\noindent Our contributions are threefold:
\begin{itemize} 
    \item We design a vibrating event camera using a rotating unbalanced mass that enables event generation in static scenes, addressing a key limitation of event cameras. 
    \item We introduce a real-time motion-compensation pipeline that estimates and removes induced vibrations online, without requiring calibration or prior knowledge of physical parameters.
    \item We validate our method on four real-world datasets, demonstrating improved event density and higher-quality results for edge detection and image reconstruction. We also show extensions to specialized applications such as scene frequency estimation and relative depth prediction. 
\end{itemize}

%% file: figs/qualitative_eval_method.tex
\begin{figure*}[ht!]
\centering
\includegraphics[width=0.975\linewidth]{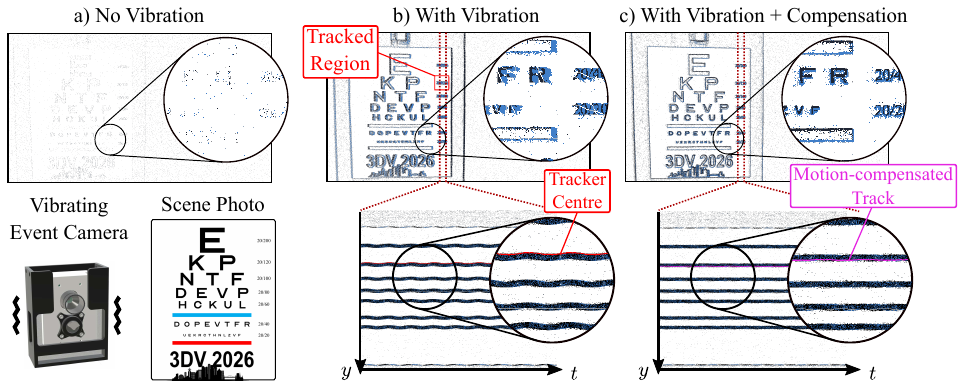}
\vspace{-5px}
\caption{\textbf{Qualitative illustration of our method.}  
\textbf{(a) No Vibration.} With a static event camera and no induced motion, the accumulated event image appears blurred and lacks sharp edges, while the $y$–$t$ slice shows little temporal structure.  
\textbf{(b) With Vibration.} Introducing controlled vibrations stimulates the sensor, increasing the number of events and producing sinusoidal traces in the $y$–$t$ slice. We develop an Extended Kalman Filter (EKF) to track the vibrational motion in a region of the scene (red).  
\textbf{(c) With Vibration + Compensation.} We use the EKF-tracked region to estimate and compensate for the sinusoidal motion across the entire scene. The motion-compensated accumulated image recovers sharp structures, and the $y$–$t$ slice aligns with the stable motion-compensated track (magenta), revealing the underlying scene.  
An inset (bottom left) shows the reference scene displayed in front of the camera with induced motion.}\label{fig:qualitative_eval_method}
\vspace{-3.2ex}
\end{figure*}

%% file: sections/02_related.tex
\section{Related Work} 
\label{sec:related}
Event cameras are increasingly used in robotics and computer vision due to their high temporal resolution and ability to capture fast scene dynamics. In this section, we review their role in perception, outline complementary hybrid sensing strategies, and discuss microsaccade-inspired methods for maintaining event generation in otherwise static scenes.

\paragraph{Event cameras in robotics.}
Event cameras offer several advantages over conventional cameras, including high dynamic range and microsecond-level temporal resolution. These properties have been exploited for feature tracking under challenging conditions~\cite{gehrig2020eklt, messikommer2023data, Zhang_2022_CVPR}.

In robotics, event cameras have been applied to visual localization~\cite{Rosinol_2018_RAL}, edge detection~\cite{brandli2016EBCCSP}, and motion compensation~\cite{Gallego2018CVPR}, frequency estimation~\cite{pfrommer2022frequency,bane2024non,spetlik2025efficient}, image reconstruction~\cite{Rosinol_2018_RAL,Jing21CVPR,guo2023sensors} and deblurring~\cite{pan2019CVPR, jiang2020CVPR}. We show that for motion compensation, edge detection, and image reconstruction tasks, our method can provide a high-quality information stream, taking advantage of the induced motion.

\paragraph{Hybrid-modality sensing methods.}
Beyond microsaccades, other strategies for event generation include moving stereo event cameras at high speed for 3D reconstruction~\cite{li20243d}, pairing event cameras with conventional frames for static localization~\cite{dong2021standard}, and employing visual tracking for object following~\cite{glover2017robust}. Machine learning-based approaches have also been developed to infer object motion from sparse event data~\cite{chen2022ecsnet}, with some methods relying on continuous-motion assumptions, for example in human choreography capture applications~\cite{xu2020eventcap}. 
Our system does not rely on conventional image frames or on the assumption of fast motions in the scene. Instead, we propose a fully model-based event-centric approach that exploits the intrinsic properties of events and their sensitivity to motion, generating a continuous stream of information. This stream can be leveraged for reconstruction, edge detection, and the tasks discussed above, even in otherwise static scenes.

\paragraph{Microsaccade-inspired event cameras.}
Inspired by fixational eye movements in human vision, several works have proposed artificial microsaccades to enable event generation in static scenes. Existing solutions rely on mechanically induced motion using pan--tilt units~\cite{yousefzadeh2018pantilt,testa2020dynamic,testa2023active,Orchard15fns,d2025wandering}, mirrors~\cite{lohr2018contrast}, rotating wedge prisms~\cite{botao2024microsaccade}, or polarization-based optical elements~\cite{maeda2025event}. These systems typically require specialized hardware and careful calibration.
Our approach employs a simpler mechanical setup and a processing pipeline that adapts automatically to system parameters, making it more suitable for real-world deployment. Concurrent work has explored motion recovery from unstructured camera jitter inspired by spacecraft vibration~\cite{bagchi2025event}, but the lack of a motion prior limits reconstruction accuracy.

The approach most similar to our own is that of He et al.~\cite{botao2024microsaccade}. They employ a rotating wedge prism to generate events continuously and utilize position-encoding hardware to track and remove the induced motion. %
Our method introduces a mechanically simpler approach, requiring no custom optical elements or position encoders such as in~\cite{botao2024microsaccade}. Instead, we directly estimate the state of the sinusoidal motion from the event stream itself and perform parameter estimation online.
Moreover, we demonstrate that our software stack is agnostic to the underlying hardware. On data generated by the rotating wedge-prism in~\cite{botao2024microsaccade}, we show that our system performs equally well in terms of motion compensation quality and event stream consistency.
Finally, our experiments show that our system enables the same downstream use cases proposed in ~\cite{botao2024microsaccade}---particularly in image reconstruction and edges extraction---with simpler hardware and a more general motion-tracking framework.

%% file: sections/03_method.tex
\section{Methodology}
\label{sec:method}
Our approach aims to excite the pixels of an event camera to produce a continuous stream of events, enabling high-quality image reconstruction and sharp edge detection from raw event data. %
We first describe the mechanical system responsible for generating vibrational motion and its effect on event generation in the camera~(\autoref{sec:motion_model}). We then present the motion compensation pipeline that allows us to recover the underlying scene structure~(\autoref{sec:params_estimation}). Finally, we describe our physical hardware prototype~(\autoref{sec:hardware_proto}).

\subsection{Camera Motion Model}
\label{sec:motion_model}
\input{figs/mass_spring_damper}
\input{figs/camera_model}
To characterize the motion induced by our vibration mechanism, we adopt a classical mass-spring-damper model, illustrated in~\autoref{fig:mass_spring_damper_figure}. This model captures the essential dynamics of the periodic motion imparted to the camera.

We model our system as a forced damped harmonic oscillator, which can be described by a second-order differential equation. The closed-form steady-state solution is of the form
\vspace*{-3mm}
\begin{align}
y(t)=\hat{A}\cdot \sin(\hat{\omega} t-\phi),
\end{align}

\noindent where $\hat{A}=\smash{\sqrt{\hat{A}_X^2+\hat{A}_Y^2}}$ is the oscillation amplitude, with $\hat{A}_X$ and $\hat{A}_Y$ as the components along $\mathbf{x}$- and $\mathbf{y}$-axis in camera coordinates. As such, the motion of the camera when viewing a static scene follows a sinusoid. %
Further details are provided in the supplementary material, \autoref{app:cam_motion_details}.

The camera motion is described by introducing a \textit{virtual camera frame}, denoted as $\camvir$, shown in~\autoref{fig:camera_model}. In our formulation, $\camvir$ remains static and acts as the reference, while the physical camera, $\cam$, moves along a controlled, circular trajectory in the $x$--$y$ image plane of $\camvir$. %

We adopt a standard pinhole projection model to describe how a 3D point $\pointw$ in the world frame $\wframe$ projects onto the image plane of $\camvir$. We denote $\pointcplanevir=(u, v, 1)$ as the projected point in the virtual frame, which is the point we would like to recover. The moving camera observes point $\pointcplane=(u',v', 1)$. The transformation from $\camvir$ to the moving camera $\cam$ is defined as %
\begin{align}
    \tccvir = 
    \begin{bmatrix}
        \mathbf{I} & 
        \begin{array}{c}
        \hat{A}_X \cos(\omegahat t + \phi_X) \\
        \hat{A}_Y \cos(\omegahat t + \phi_Y) \\
        0 
        \end{array} \\
        \mathbf{0}^\top & 1
    \end{bmatrix},
\end{align}
where $\mathbf{I}$ is the $3\times3$ identity matrix, 
$\omegahat$ is the angular frequency of the motion, and $\phi_X$, $\phi_Y$ are the corresponding phase shifts.  Note that $\omegahat$ differs from $\omega$ in~\autoref{fig:camera_model} due to inertial factors. The $\mathbf{z}$-coordinate remains constant, as the motion is planar. %
The projection of $\pointw$ onto the moving camera $\cam$ is then
\begin{align}
    \pointcplane = \mathbf{K} \tccvir \tcvirw \pointw. \label{eq:camera_model}
\end{align}
Here, $\mathbf{K}$ is the intrinsic matrix and $\tcvirw$ is the extrinsic transform that expresses $\pointw$ in $\camvir$. To illustrate this relation more concretely, consider the vertical coordinate $v$ of the projected point $\pointcplanevir$. In the $x$--$y$ image plane, the point $\pointcplanevir$ travels along a circular path with amplitude $A=\sqrt{A_x^2+A_y^2}$. Then expanding~\autoref{eq:camera_model} we obtain
\begin{align}
    v' = f\frac{Y}{Z} + f\frac{\hat{A}_Y \cos(\omegahat t + \phi_Y)}{Z} + c_Y,
\end{align}
where the term, $f\frac{Y}{Z} + c_Y$, corresponds to the projection $v$ of $\pointw$ in the static virtual frame $\camvir$, and the second term models the time-varying displacement due to camera motion. We rewrite this as %
\begin{align}
    v' = v + A_y \cos(\omegahat t + \phi_Y), \label{eq:vprime_proj_final}
\end{align}
where $A_y = f \frac{Y_0}{Z}$ is the amplitude of the induced sinusoidal motion as in~\autoref{fig:camera_model}. 

From~\autoref{eq:vprime_proj_final}, we identify three key unknowns for motion compensation along one axis: the amplitude $A_y$ (dependent on scene depth and oscillation magnitude), the phase $\phi_Y$, and the coordinate $v$ in the virtual frame. Accurate estimation of these parameters allows precise inversion of the apparent motion in the event stream and recovery of the underlying scene as if captured by a stationary camera.
\subsection{Motion Compensation Setup}
\label{sec:params_estimation}
\input{figs/method}
Once a model of the induced motion has been established, the next step is to eliminate motion from the incoming event stream. This requires estimating the dynamic motion parameters, which vary as the camera or the scene changes over time. To achieve this, we employ a series of software modules, illustrated in~\autoref{fig:method_schematic}. Below, we describe each stage of the pipeline and its role in accurate motion compensation.
\subsubsection{Event Tracker}
~\label{sec:tracker}
 We first extract $N$ centroid trajectories ${(u'_i, v'_i, t_i)}_{i=1}^N$ using an event tracker. In this work, we use HASTE~\cite{alzugaray2020haste}, a lightweight algorithm that operates directly on the raw event stream. HASTE tracks local spatiotemporal patterns without requiring additional image processing, providing robust estimates of centroid positions over time. These trajectories capture the periodic motion induced by the camera and are subsequently used to estimate the unknown parameters of~\autoref{eq:vprime_proj_final}. To initialize the HASTE trackers, we manually select regions with high texture or edge content.
 
\subsubsection{Frequency Estimation with NUFFT}
\label{sec:nufft}
Since the camera motion is oscillatory, recovering its frequency is essential. However, events are sampled irregularly in time, which makes standard Fourier analysis unsuitable.
To obtain an initial estimate of the angular velocity parameter in~\autoref{eq:vprime_proj_final}, we employ the non-uniform fast Fourier transform (NUFFT)~\cite{barnett2019SIAM}. The NUFFT decomposes a signal into its constituent frequencies but is specifically designed to operate on irregularly sampled data---making it ideal for event-based inputs that are inherently asynchronous and non-uniform in time.
Before applying the NUFFT, we remove the DC component from the tracker signal~(\autoref{sec:tracker}) to isolate oscillatory motion from the static offset introduced by the patch location. Temporal samples $t_i$ are then rescaled to $[-\pi, \pi]$, yielding normalized timestamps $\bar{t}_i$.
This produces a zero-mean, time-normalized samples ${(\bar{u'}_i, \bar{v'}_i, \bar{t}_i)}_{i=1}^{N}$ which are passed to the NUFFT. The transform returns the top $M$ dominant frequencies, from which we extract the angular frequency $\omegahat$ of the camera motion. Estimation is performed independently for the $X$ and $Y$ axes.
Using $\omegahat$, we fit the tracked samples to a sinusoidal model using least-squares optimization, recovering the amplitudes and phase offsets of the motion. These parameters initialize the extended Kalman filter (EKF) described in~\autoref{sec:ekf}.
\subsubsection{Extended Kalman Filter Setup}
\label{sec:ekf}

 An extended Kalman filter (EKF) refines the motion parameters over time by incorporating a dynamical model and accounting for observation noise. This yields temporally consistent parameter estimates, which are essential for reliable motion compensation. We model all motion parameters dynamically in the EKF, but they are updated differently. Oscillatory frequency and phase are assumed to be globally constant across the image plane, whereas the oscillation amplitude depends on the scene depth. We make the assumption that all objects in the same scene share the same depth plane, although our implementation allows for multiple trackers which may have regions with different depths. Thus, we vary the $u',v'$ parameters with each specific tracker to allow for tracking at different depths.

The camera frame's angular position $\theta$ evolves as
\begin{align}
\theta_t = \theta_{t-1} + \omegahat \Delta t,
\end{align}
where $\Delta t$ is the time elapsed since the last update. The projected event position in the $Y$ direction is modeled as
\begin{align}
v' = a_y \sin(\theta) + b_y \cos(\theta) + v, \label{eq:ekf_motion_model}
\end{align}
which is equivalent to~\autoref{eq:vprime_proj_final} but avoids the numerical instability that arises from directly modeling the phase shift $\phi$ or explicit time dependence. This formulation also mitigates ambiguities caused by trigonometric wrapping. 
The parameters for the $y$-axis expression in~\autoref{eq:vprime_proj_final} can be recovered as
\begin{align}
A_Y = \sqrt{a_y^2 + b_y^2}, \
\phi_Y = \arctan2(b_y, a_y).
\end{align}

The EKF state vector is defined as
\begin{align}
\mathbf{x} = \begin{bmatrix}
\theta & \omegahat & a_y & b_y & v \label{eq:ekf_state}
\end{bmatrix}^\top.
\end{align}
This state is updated as new tracker measurements arrive, with the corresponding Jacobians provided in Supplementary~\autoref{app:ekf_appendix}. 
The filter can operate on the entire image plane or, more effectively, on localized patches centered around feature points. Our implementation allows multiple trackers to be instantiated in parallel, enabling flexible motion parameter estimation across the scene.

The EKF enables us to predict the expected oscillatory amplitude of the incoming event stream at any given time, and subsequently remove it to obtain a motion-compensated event stream, as illustrated in panel (c) of~\autoref{fig:qualitative_eval_method}.

\subsection{Hardware Prototype}
\label{sec:hardware_proto}
For all real-world data collection and experiments, we use a Prophesee EVK3 event camera equipped with an IMX636 sensor (1280 × 720 pixels). To induce motion, we rigidly attach a DC motor~\cite{hobby_motor_sparkfun} to the camera body and mount an off-centre mass on the motor shaft. The mass is custom-machined; however, any off-axis weight with sufficient eccentricity and mass is sufficient to generate the desired motion.

The event camera is enclosed in a 3D-printed casing wrapped in foam, which acts as a passive spring–damper system and enables the motion model described in~\autoref{sec:motion_model}. The case constrains the motion to a 2D plane and provides a rigid interface for real-world mounting. Additional details on the mechanical prototype are provided in the supplementary material (\autoref{app:mechanical_setup}). %

%% file: figs/mass_spring_damper.tex
\begin{figure}[!t]
\centering
\includegraphics[width=\linewidth]{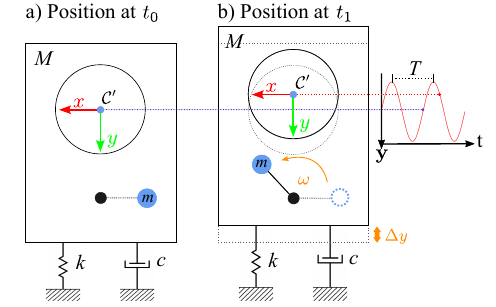}
\vspace{-4ex}
\caption{\textbf{Schematic of the mass–spring–damper model}. Camera setup at two time steps, (a) $t_0$ and (b) $t_1$. The rotation of an off-axis mass $m$ with angular velocity $\omega$ induces planar displacements $\Delta x, \Delta y$ (only the vertical component is illustrated). The estimated oscillation frequency, $\frac{2\pi}{T}$, corresponds to the motion perceived by the camera. Note that this differs from the natural frequency of the off-axis mass $\omega$, due to inertial effects. 
}
\vspace{1ex}
\label{fig:mass_spring_damper_figure}
\end{figure}

%% file: figs/camera_model.tex
\begin{figure}[t!]
\centering
\includegraphics[width=\linewidth]{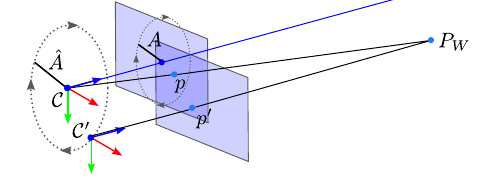}
\vspace{-3ex}
\caption{\textbf{Visual representation of the camera model}. The virtual camera $\camvir$ remains static, while the real camera $\cam$ translates along a circular path centered at $\camvir$ in the $x$–$y$ image plane of $\camvir$. We denote $\pointcplanevir$ and $\pointcplane$ as the projected point of $\pointw$ in the virtual and and real image planes respectively. $\hat{A}$ and $A$ refer to the amplitudes of motion in the virtual and real frames respectively.
The objective is to remove the resulting oscillatory motion and recover the projection of point $P_W$ onto the image plane of $\camvir$. 
}
\label{fig:camera_model}
\vspace{-2ex}
\end{figure}

%% file: figs/method.tex
\begin{figure}[!t]
    \centering
    \includegraphics[width=\linewidth]{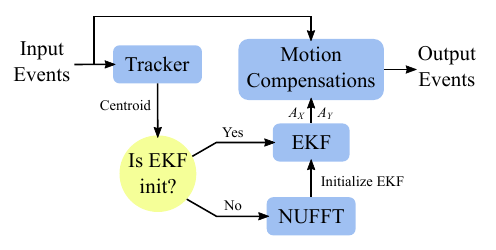} \hfill
    \vspace{-5ex}
    \caption{\textbf{Schematic representation of~\methodname}.  
 The input event stream is processed by a tracker that extracts trajectories used to estimate the dominant frequency, amplitude, and phase shift of the oscillatory motion. Once the sinusoidal motion is characterized, an Extended Kalman Filter (EKF) is initialized independently for each axis to track the induced motion. The EKF estimates are then used to compensate for the induced motion in the incoming event stream. %
 }
\label{fig:method_schematic}
\vspace{-2ex}
\end{figure}

%% file: sections/04_experiments.tex
\section{Results}
\label{sec:results}

\input{figs/results/niqe_entropy}
We evaluated \methodname\ on three real-world scenes captured with the \prophesee EVK3 (1280x720 px), demonstrating efficient motion-compensated event generation while preserving reconstruction quality and edge detail. Additional validation on the AMI-EV dataset~\cite{botao2024microsaccade}, recorded with a DVXplorer (640×480 px), confirms \methodname effectiveness and hardware independence.

\subsection{Data Collection}
We recorded real-world scenes and produced output data for three settings. 
First, we captured the scene using a standard event camera (S-EV). 
Second, we capture the scene with vibration (V-EV). 
Finally, we apply our motion compensation algorithm to the acquired V-EV data (\methodnameshort).
Overall, we evaluated our method on four real-world scenes: \textit{AMI-EV}~\cite{botao2024microsaccade}, \textit{Logo}, \textit{Pattern Checkerboard}, and \textit{Pattern}. Each scene consists of a textured pattern moving in front of the camera. The \textit{Logo} pattern contains text and rich textures, moving slowly while performing partial rotations. The \textit{Pattern Checkerboard} and \textit{Pattern} sequences involve planar patterns moving at different speeds, with the \textit{Pattern} exhibiting faster motions. Images of all patterns are provided in Supplementary~\autoref{app:dataset}.
To ensure a fair comparison in the real-world setting, we employed a Franka robotic arm to move various patterns in front of the camera repeatably for the V-EV and S-EV settings. We mounted the camera on the physical prototype mass-spring-damper system described in~\autoref{sec:hardware_proto}. %
\subsection{Metrics}
We evaluated the performance of our method for image quality, edges, and texture detection using the following metrics. We provide mathematical definitions of all metrics in our supplementary material,~\autoref{app:metrics_equations}.

\begin{itemize}
    \item \textbf{Shannon entropy}~\cite{shannonentropy}\textbf{.} The entropy of binary images accumulated over a 10~ms window indicates the amount of information captured by the sensor. Higher values correspond to richer input data~\cite{botao2024microsaccade}.
    \item \textbf{Natural Image Quality Evaluator (NIQE)}~\cite{Mittal13spl}\textbf{.} We compute NIQE on reconstructed images from E2VID~\cite{Rebecq19cvpr, Rebecq19pami} with a 10~ms event stream window. Lower values indicate reconstructions closer to the statistical properties of natural images and higher perceptual quality.
    \item \textbf{Variance of image pixel values.} We count per-pixel events over a  10~ms window and compute the variance of the resulting image pixel values. High variance indicates well-aligned events forming sharp edges, while low variance reflects blurred or dispersed events~\cite{Gallego2018CVPR}.
    \item \textbf{Gradient magnitude of image pixel values.} We count per-pixel events over a 10~ms window and calculate the gradient magnitude of the resulting image pixel values. High values indicate strong edges and texture.
    \item \textbf{Edges continuity and fragmentation.} We count per-pixel events over a 33~ms window, smooth the resulting image with a Gaussian blur, binarize, and thin using the Zhang–Suen algorithm~\cite{ZhangSuenThinning}. From the resulting edge maps, connected components are extracted~\cite{Fiorio1996TCS, wu2005MI}, and their number reflects fragmentation (many small components imply broken or noisy edges, fewer longer ones imply better continuity). We additionally compute the average edge length~\cite{pabst2007EJIP} and count edge junctions, which together serve as indicators of edge coherence and texture reconstruction quality.
\end{itemize}

\subsection{Image Quality Evaluation}
\input{tabs/results/entropy_table}
To evaluate the consistency of the output, we compute the entropy of accumulated event frames. The plot in~\autoref{fig:entropy_niqe_comparison} reports the entropy for the \textit{Logo} scene and results for all scenes are shown in~\autoref{table:entropy_values}.
Our results show that our induced motion, although small (less than a millimeter in the $x$- and $y$-directions), is sufficient to yield a higher information gain compared to the static event camera. It is worth noting that the variance of the entropy information is low when a vibratory motion is induced, meaning that we have a consistent and stable information input that does not depend on scene dynamics~(\autoref{table:entropy_values}).
\input{tabs/results/niqe_table}

\autoref{fig:entropy_niqe_comparison} shows the NIQE and entropy results for the \textit{Logo} scene, while per-scene NIQE values are provided in~\autoref{table:niqe_values}.
The results demonstrate that induced motion consistently leads to improved reconstructions compared to a static event camera. A clear correlation emerges between entropy and image quality: higher entropy yields better NIQE scores in the S-EV setup. Conversely, when entropy is low --- such as in static or low-motion scenes --- our reconstructions remain stable and do not fade, ensuring consistent information output over time.

\subsection{Edge Extraction Evaluation}
\input{tabs/results/tab_variance_grad}
We evaluate the sharpness of the accumulated frames by computing the variance and the gradient magnitude for the three setups: S-EV, V-EV, and \methodname.~\autoref{tab:grad_var_eval} reports the average and variance of these metrics across all scenes.  

The results indicate that our method effectively compensates for induced motion, achieving higher variance and gradient magnitude values, which correspond to sharper edges and stronger structural cues.
\input{tabs/results/edges_evaluation}
We evaluate edge quality using both continuity and fragmentation metrics. In~\autoref{tab:edges_evaluation}, we report statistics for the described metrics.
The evaluation shows that \methodname produces clean, well-connected edge maps, making it well-suited for preserving object boundaries and overall shape structure. In contrast, the standard event camera without induced vibration tends to produce noisier and more fragmented edges.
\subsection{Runtime Evaluation}
Our software stack, implemented in C++ with the Metavision API by \prophesee, is evaluated by running the compensation script ten times per dataset. The average processing time is $15.28 \pm 3.7$~ns per event ($\sim65$~Mev/s), including undistortion and motion compensation.
The initial parameter estimation (\autoref{sec:nufft}), requires $104.8 \pm 0.03$~ms, but remains non-blocking thanks to a multithreaded design that tracks incoming events while buffering and performing motion compensation.
A video demonstration is provided in the supplementary material. Although runtime is fast, real-time performance depends on sensor resolution and scene dynamics; in our experiments, event rates peaked at $45$~Mev/s.

%% file: figs/results/niqe_entropy.tex
\begin{figure*}[t]
\centering
\vspace{-2.5ex}
\includegraphics[width=\linewidth]{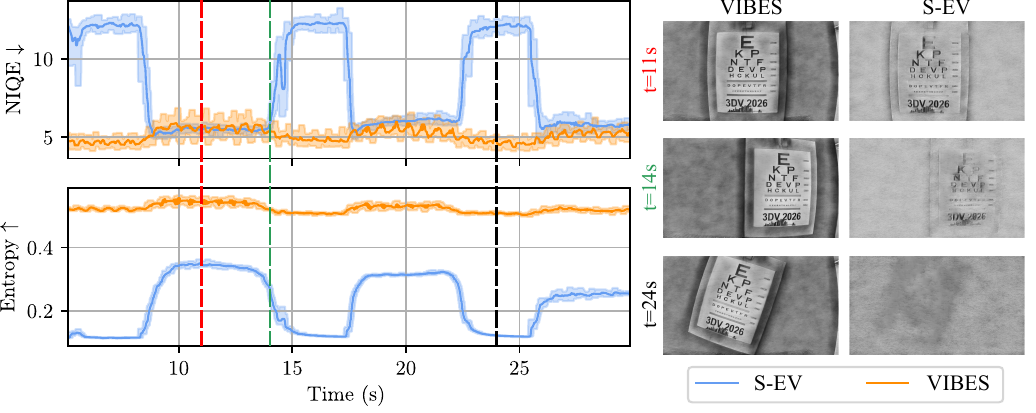}
\vspace{-5ex}
\caption{\textbf{Shannon entropy and NIQE scores} of reconstructed frames of binary accumulated event frames, computed over 10~ms time windows on the \textit{Logo} real-world scene. The shaded range denotes the minimum and maximum values within every 10-frame window. Temporal screenshots of both the \methodname and S-EV reconstructions, generated using E2VID~\cite{Rebecq19cvpr, Rebecq19pami}, are shown on the right. The regions where the S-EV method produces low-quality images are those where the scene is completely static or there is slow motion.
}
\vspace{-1ex}
\label{fig:entropy_niqe_comparison}
\end{figure*}

%% file: tabs/results/entropy_table.tex
\begin{table}[t]
\centering
\begin{tabular}{ccc}
\toprule
\textbf{Scene} & \textbf{Method} & \textbf{Entropy} $\uparrow$ \\
\midrule
\multirow{2}{*}{AMI-EV~\cite{botao2024microsaccade}} & S-EV & $0.08\pm0.05$ \\
 & \colortab \methodnameshort & \colortab $\mathbf{0.24\pm0.04}$ \\
\cmidrule{2-3}
 \multirow{2}{*}{Logo} & S-EV & $0.21\pm0.09$ \\
& \colortab \methodnameshort & \colortab $\mathbf{0.52\pm0.01}$ \\
\cmidrule{2-3}
 \multirow{3}{*}{\makecell{Pattern\\Checkerboard}} & S-EV & $0.29\pm0.13$ \\
& \colortab \methodnameshort & \colortab $\mathbf{0.59\pm0.04}$ \\
\cmidrule{2-3}
 \multirow{2}{*}{Pattern} & S-EV & $0.30\pm0.12$ \\
 & \colortab \methodnameshort & \colortab $\mathbf{0.39\pm0.08}$ \\
\bottomrule
\end{tabular}
\caption{\textbf{Shannon entropy} values (mean, standard deviation) computed on binary accumulated event frames over 10~ms windows. Higher entropy indicates a richer and more informative event stream. %
}
\vspace{-2ex}
\label{table:entropy_values}
\end{table}

%% file: tabs/results/niqe_table.tex
\begin{table}[t]
\centering
\begin{tabular}{ccc}
\toprule
\textbf{Scene} & \textbf{Method} & \textbf{NIQE} $\downarrow$\\
\midrule
     \multirow{2}{*}{AMI-EV~\cite{botao2024microsaccade}} & S-EV & $10.4 \pm 5.0$ \\
 & \colortab V-EV & \colortab $\mathbf{9.1} \pm  1.0$ \\
\cmidrule{2-3}
 \multirow{2}{*}{Logo} & S-EV & $8.6 \pm 9.4$ \\
 & \colortab V-EV & \colortab $\mathbf{5.1} \pm 0.3$ \\
\cmidrule{2-3}
 \multirow{2}{*}{\makecell{Pattern\\Checkerboard}} & S-EV & $8.2 \pm 9.0$ \\
  & \colortab V-EV & \colortab $\mathbf{6.3} \pm 0.2$ \\
\cmidrule{2-3}
 \multirow{2}{*}{Pattern}& S-EV & $9.4 \pm 5.0$ \\
 & \colortab V-EV & \colortab $\mathbf{8.2} \pm 0.6$ \\
\bottomrule
\end{tabular}
\caption{\textbf{Natural Image Quality Evaluator (NIQE)} average values (mean, standard deviation) computed using E2VID~\cite{Rebecq19cvpr} at 100 Hz.}
\vspace{-3ex}
\label{table:niqe_values}
\end{table}

%% file: tabs/results/tab_variance_grad.tex
\begin{table}[t]
\centering
\begin{tabular}{cccc}
\toprule
\textbf{Scene} & \textbf{Method} & \textbf{Variance} $\uparrow$  & \textbf{Grad. Mag.} $\uparrow$ \\
\midrule
\multirow{3}{*}{AMI-EV~\cite{botao2024microsaccade}} & S-EV & $0.06 \pm 0.05$ & $0.04 \pm 0.02$ \\
 & \colortab V-EV & \colortab $0.32 \pm 0.06$ & \colortab $0.09 \pm 0.02$ \\
 & \methodnameshort & $\mathbf{0.44 \pm 0.10}$ & $\mathbf{0.12 \pm 0.02}$ \\
\cmidrule{2-4}
 \multirow{3}{*}{Logo} & S-EV & $0.11 \pm 0.08$ & $0.18 \pm 0.07$ \\
 & \colortab V-EV & \colortab $2.34 \pm 0.13$ & \colortab $0.48 \pm 0.03$\\
 & \methodnameshort & $\mathbf{3.86 \pm 0.34}$ & $\mathbf{0.56 \pm 0.02}$ \\
\cmidrule{2-4}
 \multirow{3}{*}{\makecell{Pattern\\Checkerboard}} & S-EV & $0.16 \pm 0.11$ & $0.22 \pm 0.11$ \\
 & \colortab V-EV & \colortab $ 1.9 \pm 0.15$ & \colortab $0.57 \pm 0.04$ \\
  & \methodnameshort & $\mathbf{3.16 \pm 0.33}$ & $\mathbf{0.60 \pm 0.04}$ \\
\cmidrule{2-4}
 \multirow{3}{*}{Pattern} & S-EV & $0.22 \pm 0.18$ & $0.19 \pm 0.09$ \\
 & \colortab V-EV & \colortab $0.76 \pm 0.16$ & \colortab $\mathbf{0.30 \pm 0.06}$ \\
 & \methodnameshort & $\mathbf{1.11 \pm 0.25}$ & $\mathbf{0.30 \pm 0.06}$ \\
\bottomrule
\end{tabular}
\caption{\textbf{Quantitative evaluation of motion compensation.} We report image variance and average gradient magnitude, expressed as mean and standard deviation across frames.
Higher values indicate sharper images and better-preserved scene structure.}

\vspace{-3ex}
\label{tab:grad_var_eval}
\end{table}

%% file: tabs/results/edges_evaluation.tex
\begin{table*}[t]
\centering
\begin{tabular}{ccccc}
\toprule
\textbf{Scene} & \textbf{Method} & \textbf{Avg. Num. Components} $\downarrow$ & \textbf{Avg. Contour Length} $\uparrow$ & \textbf{Avg. Junction Count} \\
\midrule
 \multirow{2}{*}{AMI-EV~\cite{botao2024microsaccade}} & S-EV & $91.1\pm17.2$ & $4.2\pm2.4$ & $63.3\pm54.6$ \\
 & \colortab \methodname & \colortab $\mathbf{82.9\pm14.1}$ & \colortab $\mathbf{8.3\pm1.6}$ & \colortab $147.9\pm31.8$ \\
\cmidrule{2-5}
 \multirow{2}{*}{Logo} & S-EV & $265.8\pm221.4$ & $3.6\pm3.1$ & $136.5\pm165.8$ \\
 & \colortab \methodname & \colortab $\mathbf{175.0\pm52.5}$ & \colortab $\mathbf{32.1\pm10.8}$ & \colortab $979.3\pm349.2$ \\
\cmidrule{2-5}
 \multirow{2}{*}{\makecell{Pattern\\Checkerboard}} & S-EV & $300.3\pm232.9$ & $6.7\pm5.3$ & $193.2\pm176.3$ \\
  & \colortab \methodname & \colortab $\mathbf{140.7\pm40.2}$ & \colortab $\mathbf{49.9\pm17.3}$ & \colortab $691.3\pm63.2$ \\
\cmidrule{2-5}
 \multirow{2}{*}{Pattern} & S-EV & $138.7\pm102.3$ & $5.6\pm4.2$ & $226.9\pm197.5$ \\
 & \colortab \methodname & \colortab $\mathbf{116.3\pm35.1}$ & \colortab $\mathbf{15.6\pm4.2}$ & \colortab $304.4\pm127.6$ \\
\bottomrule
\end{tabular}
\caption{\textbf{Edge extraction evaluation on captured scenes.} We report the number of connected components (lower is better), the average contour length in pixels (higher is better), and the number of junctions, expressed as mean ± standard deviation across frames. Across most scenes, \methodname reduces edge fragmentation and increases contour continuity compared to S-EV, indicating cleaner, more coherent edge maps.}
\label{tab:edges_evaluation}
\vspace{-2ex}
\end{table*}

%% file: sections/05_applications.tex
\section{Applications}\label{sec:Applications}
In this section, we demonstrate two additional applications of our pipeline beyond its primary goal of showing that induced motion can generate a stable event stream suitable for high-quality image reconstruction and edge extraction.
\subsection{Frequency Estimation}
\input{tabs/results/frequency_estimation}
While our software is designed to detect the frequency of the camera’s oscillatory motion, it can also be applied to estimate the vibration frequency of an object in the scene with a static camera.

To evaluate this, we display a triangle moving along a small circular path on a computer monitor, point the event camera at the screen, and apply \methodname's initialization stack to estimate the triangle's motion frequency. Given the monitor’s refresh rate of 75 Hz, we test a range of frequencies from 7.5 Hz to 22.5 Hz, noting that higher frequencies are affected by aliasing due to temporal sampling limitations. For each target frequency, we repeat the estimation process 10 times, reporting the mean estimated frequency and the corresponding 3$\sigma$ error.

The results in~\autoref{table:freq_est} demonstrate that \methodname achieves accurate frequency estimation with low bias and variance. We note, however, a slight increase in bias at higher frequencies, with deviations of up to 0.1 Hz between the ground truth and the predicted values. This effect is likely due to aliasing, as we are able to estimate higher frequencies in both simulated data~\autoref{sec:depth_est} and real-world data.
\subsection{Relative Depth Awareness}\label{sec:depth_est}
We explore how parallax induced by the camera vibration can encode information about scene depth.
Note that this type of depth estimation is not possible in setups like the wedge-prism solution proposed by~\cite{botao2024microsaccade}, because the camera center of projection does not change. 

We test three synthetic scenes containing two staggered patterns at different depth planes, as in~\autoref{fig:depht_stagger}. To create synthetic data, we generate RGB frames using Blender~\cite{blender} at 1000~FPS and convert them to events using V2E~\cite{delbruck2020v2e}.
Camera motion is simulated according to the sinusoidal motion model described in~\autoref{sec:motion_model}. We instantiate two trackers per pattern. Then we store the amplitudes estimated by the EKF for each of the patterns and average them across the entire scene duration. 
Finally, we compute the ratio between the average amplitudes to get the estimation of the relative depth. 
Our simulated results show that our system reliably predicts the relative depth of the two patterns: for ground-truth relative depth ratios of $0.33$, $0.50$, and $0.66$ the predicted ratios are $0.41 \pm 0.01$, $0.59 \pm 0.02$, and $0.80 \pm 0.03$.
Please see Supplementary~\autoref{app:depth_est} for additional details.

\input{figs/stagger_depth_representation}

%% file: tabs/results/frequency_estimation.tex
\begin{table}[t]
\centering
\begin{tabular}{c|ccc}
\toprule
\textbf{Target (Hz)} & \textbf{Est. (Hz)} & \textbf{Avg. Abs. Error} $\downarrow$ & \textbf{3$\boldsymbol{\sigma}$} $\downarrow$ \\
\midrule
7.5   & 7.486  & 0.0617 & 0.202 \\
\colorrow 10.0  & 10.074 & 0.0754 & 0.177 \\
12.6  & 12.684 & 0.0923 & 0.234 \\
\colorrow 15.2  & 15.255 & 0.0764 & 0.264 \\
17.4  & 17.476 & 0.0829 & 0.225 \\
\colorrow 20.0  & 20.100 & 0.1000 & 0.189 \\
22.6  & 22.644 & 0.0517 & 0.147 \\
\bottomrule
\end{tabular}
\caption{\textbf{Scene frequency estimation}. \methodname performs precise estimation of frequency, although performance degrades as the frequency increases. Reported results are based on 10 samples.}
\vspace{-3ex}
\label{table:freq_est}
\end{table}

%% file: figs/stagger_depth_representation.tex
\begin{figure}[t!]
\centering
\includegraphics[width=\linewidth]{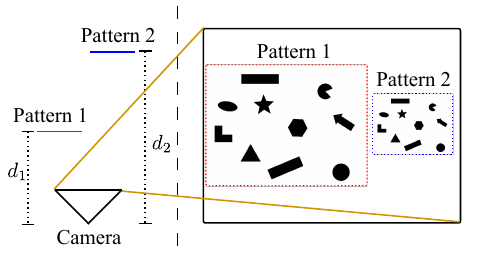}
\vspace{-3ex}
\caption{\textbf{Illustration of depth prediction setup}. The two patterns are placed at different distances from the camera and we estimate the ratio $\frac{d_1}{d_2}$ (or vice versa) by observing the amplitudes of the perceived motion of the camera measured by the tracker~(\autoref{sec:tracker}).}
\label{fig:depht_stagger}
\vspace{-3ex}
\end{figure}

%% file: sections/06_conclusion.tex
\section{Conclusion}
\label{sec:conclusion}

In this work, we introduced \methodname, a bio-inspired approach that enhances event cameras by actively inducing motion.
Our method generates persistent event streams in static scenes, addressing a key limitation of event-based vision.

We see several promising avenues for future work.
For example, the vibration system could be made adaptive---that is, disabled in highly dynamic scenes with dense event streams, and reactivated when the event density drops. Such an approach would preserve the low-power benefits of event cameras while ensuring a continuous visual signal when needed. 
Additionally, although our method already handles high event rates, downsampling or other event filtering techniques may be considered to further improve the real time performance, given the high event bandwidth generated by the induced motion. 
Our approach could also be combined with other event-based algorithms to improve performance on downstream tasks.

%% file: sections/07_supplementary.tex
\renewcommand{\thesection}{S\arabic{section}}
\renewcommand{\thetable}{S\arabic{table}}
\renewcommand{\thefigure}{S\arabic{figure}}
\renewcommand{\theequation}{S\arabic{equation}}

\setcounter{section}{0}
\setcounter{table}{0}
\setcounter{figure}{0}
\setcounter{equation}{0}

\section{Supplementary Material}
\label{app:appendix}

\subsection{Motion Model Details}
\label{app:cam_motion_details}
Without loss of generality, we focus on one axis, $y$, noting that the motion along the $x$ axis is equivalent up to a phase shift. The displacement of the camera as a function of time follows the steady-state solution of a damped harmonic oscillator, governed by the second-order differential equation
\begin{align}
M\ddot{y}(t) + c\dot{y}(t) + ky(t) =me\omega^2 \cos(\omega t),
\end{align}
where $M$ is the total system mass, $m$ is the eccentric mass, $e$ is the eccentricity, $c$ is the damping coefficient, and $k$ is the spring constant. The external force driving the system has amplitude $F_0=me\omega^2$ and angular frequency $\omega$. The exact angular frequency generated by the motor is modeled in~\autoref{app:motor_modelling}. $y(t)$, $\dot{y}(t)$ and $\ddot{y}(t)$ denotes the displacement, velocity and acceleration, of the camera respectively.
After transient dynamics have decayed, the system reaches a steady-state oscillation described by
\begin{align}
\hat{A}_Y &= \frac{F_0}{\sqrt{(k - m\omega^2)^2 + (c\omega)^2}}, \\
\phi &= \tan^{-1}\left(\frac{c\omega}{k - m\omega^2}\right), \\
x_s(t) &= \hat{A}_Y \sin    (\omega t - \phi). \label{eq:motion_eq}
\end{align}
This solution describes a smooth, predictable sinusoidal motion that ensures events are continuously generated, even when viewing a static scene. Since the motion is mechanically induced and follows a known trajectory, it provides a strong prior for both data association and noise filtering in subsequent stages of our pipeline.

\subsubsection{Motor Modelling}
\label{app:motor_modelling}
\input{tabs/exp_motor_values}

By varying the applied voltage to the motor, we can control the speed and, consequently, the amplitude and frequency of the induced harmonic motion. To relate the applied voltage $V_A$ to the motor’s angular speed $\omega_m$, we use a classical DC motor model, depicted in Figure~\ref{fig:motor_model}:
\input{figs/motor_model}

\noindent
where $i_A$ is the armature current, $k_\phi$ is the motor constant, and $R$ is the armature resistance. The back electromotive force (EMF) $E_a$ is proportional to the rotational speed $\omega_m$. Neglecting inductance (which has minimal impact in steady-state operation), the motor equations are
\begin{align}
    V_A &= i_A R + E_a = i_A R + k_\phi \omega_m \, , \\
    \omega_m &= \frac{V_A}{k_\phi} - \frac{R}{(k_\phi)^2} T_q \, ,
\end{align}
We estimate the motor constant $k_\phi$ using manufacturer specifications.~\autoref{tab:motor_values} shows a full list of motor parameters. Specifically, we calculate $k_\phi$ using 
\[ \omega_n^N=\frac{V_A^N}{k_\phi}, \]
where values superscripted with $N$ indicate nominal or rated values, as reported by the manufacturer, and are shown in~\autoref{tab:motor_values}. Here, $T_q$ is the load torque at the motor shaft, determined by the weight and eccentricity of the eccentric mass. We approximate $T_q$ using the following
\[ T_q = e\cdot m\cdot g.\]
\subsubsection{MATLAB Modelling}

Vibrations were induced by applying a voltage of 2.0~V to the DC motor setup, producing angular velocities in the range $(250, 350)$~rad/s. More details are provided in~\autoref{app:mechanical_setup} of our supplementary material.

For frequency estimation, the NUFFT initialization parameters were set to $\omega_{\text{min}} = 30$~rad/s and $\omega_{\text{max}} = 500$~rad/s, matching the expected frequency range of the mechanical system.

\section{Extended Kalman Filter Details}
\label{app:ekf_appendix}
For the EKF, we require a compact parametrization of the oscillatory motion that is numerically stable and avoids explicit dependence on absolute time $t$. We model the vertical image coordinate $v'(t_k)$ (an analogous formulation applies to the horizontal coordinate) in discrete time as

\begin{align}
v'_k 
  &= A_{y, k} \cos(\omegahat_k + \phi_Y) + v_k \\
  &= A_{y, k} \big[\cos(\omegahat_k)\cos\phi_Y \nonumber\\ & \quad\quad - \sin(\omegahat t_k)\sin\phi_Y \big] + v_k \\
  &= \underbrace{(A_{y, k} \cos\phi_Y)}_{b_{y, k}}\, \cos(\omegahat_k) 
   \nonumber\\ & \quad\quad + \underbrace{(-A_{y, k} \sin\phi_Y)}_{a_{y, k}} \, \sin(\omegahat_k) + v_k \\
  &= a_{y, k} \sin(\theta_k) + b_y \cos(\theta_k) + v_k, \label{eq:state_transition} 
\end{align}
where we augment the state with $\theta_k$ defined as
\begin{align}
\theta_k = \theta_{k-1} + \omegahat_k \Delta t, \label{eq:propagation_eq}
\end{align}
with $\Delta t = t_k - t_{k-1}$ the elapsed time between steps. 

This reparameterization introduces the coefficients $(a_y, b_y)$ instead of $(A_{y, k}, \phi_Y)$, which has several advantages:  
(i) it avoids the explicit trigonometric wrapping of the phase $\phi_Y$,  
(ii) it ensures smooth Jacobians for the EKF update step, and  
(iii) it makes the model linear in $(a_y, b_y, v)$, with only $\theta$ evolving nonlinearly.  

The same formulation applies to the horizontal coordinate $u'(t_k)$ with parameters $(a_x, b_x)$. Without loss of generality, and to simplify notation, we drop the $x$/$y$ subscripts and use a generic $c$ to indicate the DC component of the sinusoid. The complete EKF state vector for the sinusoid is then
\begin{align}
\mathbf{x}_k = 
\begin{bmatrix}
\theta_k & \omega_k & a_k & b_k & c_k
\end{bmatrix}^T,
\end{align}
where $c_k$ denotes the motion-compensated, stable image coordinate in the virtual frame. 
The state covariance is
$\mathbf{P}_k \in \mathbb{R}^{5 \times 5}$.

\subsection{Propagation Step}
The discrete-time process model is
\begin{equation}
\mathbf{x}_{k|k-1} =
\begin{bmatrix}
\theta_{k-1} + \omega_{k-1} \Delta t \\
\omega_{k-1} \\
a_{k-1} \\
b_{k-1} \\
c_{k-1}
\end{bmatrix}.
\end{equation}

Given the propagation~\autoref{eq:propagation_eq}, the state transition Jacobian is
\begin{equation}
\mathbf{F}_k =
\frac{\partial f}{\partial \mathbf{x}} =
\begin{bmatrix}
1 & \Delta t & 0 & 0 & 0 \\
0 & 1 & 0 & 0 & 0 \\
0 & 0 & 1 & 0 & 0 \\
0 & 0 & 0 & 1 & 0 \\
0 & 0 & 0 & 0 & 1
\end{bmatrix}.
\end{equation}

The predicted covariance is updated as
\begin{equation}
\mathbf{P}_{k|k-1} = \mathbf{F}_k \, \mathbf{P}_{k-1} \, \mathbf{F}_k^\top + \mathbf{Q},
\end{equation}
where $\mathbf{Q} \in \mathbb{R}^{5 \times 5}$ is the process noise covariance.

\subsection{Measurement Update}
The observation model at time $t_k$ is given by~\autoref{eq:state_transition}, with measurement noise $n_k \sim \mathcal{N}(0, \sigma_r^2)$:
\begin{align}
    h_k = a_k \sin(\theta_k) + b_k \cos(\theta_k) + c_k + n_k.
\end{align}
The measurement Jacobian is
\begin{align}
\mathbf{H}_k &=
\frac{\partial h_k}{\partial \mathbf{x}} \\
&=
\begin{bmatrix}
a_k \cos(\theta_k) - b_k \sin(\theta_k) & 0 & \sin(\theta_k) & \cos(\theta_k) & 1
\end{bmatrix}.
\end{align}
The innovation is
\begin{equation}
\tilde{y}_k = y_k - h(\mathbf{x}_{k|k-1}),
\end{equation}
with innovation covariance
\begin{equation}
S_k = \mathbf{H}_k \, \mathbf{P}_{k|k-1} \, \mathbf{H}_k^\top + \mathbf{R}.
\end{equation}
where $\mathbf{R}$ is the measurement noise covariance. 

The Kalman gain is
\begin{equation}
\mathbf{K}_k = \mathbf{P}_{k|k-1} \, \mathbf{H}_k^\top \, S_k^{-1}.
\end{equation}

The state and covariance are updated using the Joseph form
\begin{align}
\mathbf{x}_k &= \mathbf{x}_{k|k-1} + \mathbf{K}_k \tilde{y}_k, \\
\mathbf{P}_k &= (\mathbf{I} - \mathbf{K}_k \mathbf{H}_k) \,
\mathbf{P}_{k|k-1} \,
(\mathbf{I} - \mathbf{K}_k \mathbf{H}_k)^\top
+ \mathbf{K}_k \, \sigma_r^2 \, \mathbf{K}_k^\top.
\end{align}
This form preserves numerical stability and guarantees that $\mathbf{P}_k$ remains positive semi-definite.

\subsection{Physical Prototype Details}
\label{app:mechanical_setup}
\input{figs/mech_system}
\input{figs/real_camera}
To induce motion in the camera, we rigidly attach a DC motor to the body of the event camera. The motor drives an eccentric (off-axis) mass, acting as a forced rotating unbalance. This mass was custom-machined out of 1" diameter tool steel stock. The parameters of the mass are provided in~\autoref{tab:motor_values}. More detail about the modelling of the motor is provided in~\autoref{app:motor_modelling}.

An exploded view of the entire assembly is shown in~\autoref{fig:mech_system}. All assembly components except for the event camera, and motor, are designed in SolidWorks and secured with standard fasteners. The white casing and black shell, shown in~\autoref{fig:mech_system}, are 3D printed using Polylactic Acid (PLA).

The entire camera-motor assembly is mounted to an outer shell cushioned with foam, which acts as a passive spring-damper system. This shell also limits movement of the camera in the $Z$ direction, constraining the resulting motion predominantly to the $X$-$Y$ image plane which ensures predictable, repeatable oscillations. These small oscillations stimulate the event sensor, triggering events even in otherwise static scenes and revealing fine details that would not be detectable without motion.

\subsection{Dataset} 
\label{app:dataset}

Here we provide more detailed information on the real-world dataset and synthetic datasets. 

\subsubsection{Real World Data}
\input{figs/supplementary/data_collection_env}
\input{figs/supplementary/real_world_patterns}

Our experiments were performed in a scene containing our physical VIBES prototype mounted to a table, along with a Franka Emika arm. We printed three patterns, shown in Figure~\autoref{fig:real_world_patterns}, and taped them to a sheet of cardboard, which was mounted in the gripper of the Franka arm. We then moved the patterns in front of the event camera by creating motion plans in the Franka Desk software. Figure~\autoref{fig:data_collection_env}  shows pictures of our data collection scene.

\subsubsection{Synthetic Data}
\input{tabs/supplementary/depth_stagger}
We used synthetic data generated in Blender~\cite{blender} to test the depth estimation capabilities of VIBES. 
Our virtual camera had an angular FOV of $51.7^{\circ}$ and our VIBES motion radius was set to $3$ mm. VIBES in our synthetic scene can detect differences in depth up to 148 cm (see~\autoref{app:depth_est}). Given this, we placed objects in the virtual scene well within this range to ensure depth estimation was possible. 
Table~\autoref{table:synthetic_depth_data} reports the distances of the objects in the three synthetic scenes.

\subsection{Details on Metrics Equations}
\label{app:metrics_equations}

Here we give the full mathematical definitions of the metrics used in Section~\autoref{sec:results}. 

\textbf{Shannon Entropy}~\cite{shannonentropy}. We calculate Shannon Entropy over binary accumulated event images, where the binary pixels are treated as drawn from the same distribution. This is calculated as follows:
\begin{align}
    H(X) := - \sum_{x\in X} p(x) \text{log}_2p(x),
\end{align}
where $x$ is each pixel in an image $\mathcal{X}$. 

\textbf{Natural Image Quality Evaluator (NIQE)}~\cite{Mittal13spl}.

We used NIQE as introduced in~\citet{Mittal13spl}. For brevity we do not include the full algorithm. Roughly, NIQE estimates natural image quality without requiring access to distorted images by calculating certain image statistics from a large dataset of natural images, and then comparing a multivariate gaussian model of those natural image statistics to the same statistics calculated on a potentially distorted image. We specifically use the MATLAB implementation which contains a pre-trained model. 

\textbf{Variance of image pixel values.} Following~\cite{Gallego2018CVPR} which found that high variance in accumulated event frames corresponds to sharper edges. We generated accumulated frames over a 10 ms window, and then calculated the variance of the per-pixel event counts:
\begin{align}
V(X) := \frac{1}{N_p} \sum_{i,j} (h_{ij} - \mu_X)^2,
\end{align}
where $N_p$ is the number of pixels of image $X$ and $\mu_X$ is the mean of $X$

\textbf{Gradient magnitude of image pixel values.} We perform the same event accumulation procedure as above, but calculate per-pixel gradients and take an average of the magnitude over the image. We approximate per-pixel gradients using forward differences which can be calculated with convolutions:
\begin{align}
G_x = \begin{bmatrix}
-1 &  1\\
\end{bmatrix} * X,
\end{align}
\begin{align}
G_y = \begin{bmatrix}
1\\
-1\\
\end{bmatrix} * X,
\end{align}
\begin{align}
M(X) =  \sqrt{G_x^2 + G_y^2}.
\end{align}
\textbf{Edges continuity and fragmentation.} 
Once the raw edge maps are smoothed and thinned, we calculate three metrics.
\begin{itemize}
\item The \textbf{average number of components} measures the number of components in an edge map, with a greater number showing less fragmented and noisy edges. We specifically consider 8-connectivity, which considers pixels to be neighbors if another pixel is in any of the 8 surrounding pixels (practically, this means regions remain connected across diagonals). 
\item The \textbf{average contour length} measures the average lengths of the edges in the edge map. Given that in our edge maps the edges are roughly $\sim1$ pixel wide, we simply calculate the area of each edge by counting the total pixels to calculate the edge length.
\item The \textbf{average junction count} measures the number of "junctions" in the edge map, which is defined as the number of pixels with more than $2$ neighbors in their 8-connectivity area (meaning in the 8 pixels surrounding a pixel, at least 2 are occupied by other positive edge pixels). 
\end{itemize}

\subsection{Limits of Depth Estimation under Camera-Induced Motion}
\label{app:depth_est}

Given the pinhole camera and the motion model, we can compute the minimum maximum distance required to generate events that cover more than one pixel, that is, after a certain distance, the induced motion can generate a stream of events with positive and negative polarity in the same pixel location.

Assume the axis of the camera's rotation is pointing at a point we wish to observe as a feature. Define the distance from the camera to the point as $d$, and the diameter of the camera's rotation as $\hat{A_y}$. We can observe that the angle from the camera's axis of rotation is given by 
\begin{align}
    \theta = \frac{\pi}{2} - \text{arctan}\left(\frac{d}{r}\right).
\end{align}
The formula for the change in pixel space of this object is given by 
\begin{align}
    \Delta p = \text{tan} (\theta) * \frac{\text{Resolution}/2}
    {\text{tan}\left(\frac{\text{FOV}}{2}\right)}.
\end{align}
We used the Prophesee EVK3 event camera with the IMX636 sensor has a resolution of ($1280$ x $720$) and a 5 mm lens focal length. Using Prophesee's optics calculator \footnote{See here: \url{https://docs.prophesee.ai/stable/hw/manuals/optics_calculator.html}} we found our camera has a vertical angular FOV of $39^{\circ}$. Our rotation radius with our gimbal setup was estimated based on a spring-dampener ODE model to be $1$mm. Using our equation, we can find the minimum value for $d$ such that $\Delta p$ is less than or equal to 1, which represents a change of less than a pixel of the feature given the camera motion. 
Solving for $d$ with an inverse solver gave a distance of $\mathbf{47.3}$ \textbf{cm} beyond which the camera cannot detect event changes through the camera motion. Due to this relatively shallow depth, we use synthetic data for our depth estimation experiment where we can easily increase the magnitude of motion to allow for deeper depth perception.
\input{tabs/supplementary/power_consumption}

\subsection{Depth Amplitude Relation}
\input{figs/supplementary/depth_amplitude}
The oscillation amplitudes estimated from the event stream inherently encode information about scene depth. As illustrated in~\autoref{fig:rel_depth_ampl}, the induced motion can be interpreted as forming a pseudo-stereo-camera model. At the motion apexes along the $y$ axis (analogously for $x$), the effective stereo baseline corresponds to the physical amplitude $\hat{A_y}$ imposed by the mass–spring–damper system.  

From stereo geometry, given a baseline $\hat{A_y}$ and disparity $\Delta v'_0 - \Delta v'_1$ --- which in our setting is observed as the oscillation amplitude $A_y$ --- the depth $d$ can be expressed as
\begin{align}
    d = f \cdot \frac{A_y}{\hat{A_y}}.
\end{align}
In practice, however, the true physical motion $\hat{A_y}$ of the camera is unknown and cannot be measured directly. Our system can only recover the apparent amplitudes $A_y$ from the incoming event stream. While this prevents absolute depth estimation, the relative scaling of amplitudes across different tracked objects still provides a strong cue to relative depth. For instance, by initializing multiple independent EKFs at different image locations and compensating events locally, one can compare their oscillation amplitudes to infer depth ordering in the scene.  

Specifically, following from~\autoref{eq:vprime_proj_final} and considering only the oscillatory component, for two points $p_1$ and $p_2$ located at depths $Z^1$ and $Z^2$, we obtain
\begin{align}
\frac{A_y^1}{A_y^2} 
    = \frac{f \cdot \tfrac{Y_0}{Z^1}}{f \cdot \tfrac{Y_0}{Z^2}} 
    = \frac{Z^2}{Z^1}, \label{eq:rel_depth}
\end{align}
where $Y_0$ denotes the (unknown but constant) physical baseline motion along the $y$ axis. This ratio directly encodes the relative distances of points from the camera.

\subsection{Power Consumption}
\label{app:power_consumption}
We report the power consumption of our rotating unbalanced mass system and compare its performance against alternative approaches. \methodname employs a DC hobby motor, with parameters listed in~\autoref{app:motor_modelling}. From our experiments, we observe that the motor draws an average current of $0.141$A, while the applied voltage was set at $2$V. Consequently, the typical power consumption of our DC motor setup is
\[ 2V\cdot 0.141A = 0.282W\]
In these experiments, the DC motor was powered via a DC/DC buck converter. We do not account for the converter’s efficiency in our calculations for two reasons. First, switched-mode DC/DC converters generally achieve high efficiency within this voltage range. Second, comparable data on power supply efficiency from other methodologies are not readily available. 

Power consumption values are summarized in~\autoref{table:power_consumption}. For each methodology, we report both the estimated power consumption and the basis of evaluation. When specific hardware is identified, we reference the corresponding manufacturer's datasheets. For example, the FLIR PTU units~\cite{flir_ptu_d46,flir_ptu_e46} list distinct power requirements for low-power and full-power modes. For \methodname, power consumption was measured directly during online experiments. For other approaches, expected values are reported, defaulting to no-load or standby power consumption when both online measurements and datasheet values are unavailable.

%% file: tabs/exp_motor_values.tex
\begin{table*}[t!]
\centering
\begin{tabular}{cccccccc}
\toprule
$\mathbf{k\phi}$ & $\mathbf{R\ [\Omega]}$ & $\mathbf{V_A^N\ [V]}$ & $\mathbf{i_A^N\ [mA]}$ & $\mathbf{\omega_m^N\ [RPM]}$ & $\mathbf{m_{mass}\ [g]}$ & $\mathbf{e\ [mm]}$ & $\mathbf{T_q\ [N\cdot mm]}$ \\
\midrule
0.00374 & 3.75 & 3 & 110 & 6600 & 7.1 & 3.1 & 0.216 \\
\bottomrule
\end{tabular}
\caption{\textbf{Motor and vibration setup parameters.} Experimental values of the motor and vibration system, including motor constants, operating parameters, and physical properties.}
\vspace{-3ex}
\label{tab:motor_values}
\end{table*}

%% file: figs/motor_model.tex
\begin{figure}[h]
\centering
\begin{circuitikz}[american voltages]
\draw   (0,0)   to[Telmech=M, n=motor] (0,1.8)
                to[R,a=$R$] ++ (-2,0)
                to[L,a=$L$] ++ (-2,0)
                to[V=$V_A$,-,i<_=$i_A$] ++ (0,-1.8)
                -- (0,0);
\node[right] at (motor.east) {$E_A=k\phi\omega_m$}; 
\end{circuitikz}
\caption{\textbf{Simple DC motor model.}}
\label{fig:motor_model}
\end{figure}
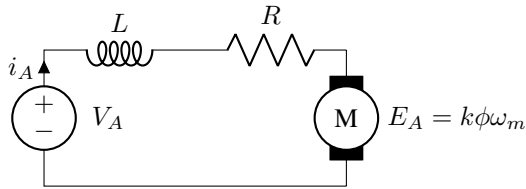

%% file: figs/mech_system.tex
\begin{figure}[t!]
\centering
  \centering
  \includegraphics[width=\linewidth]{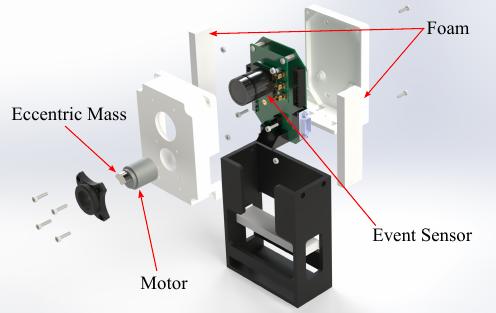}
\caption{\textbf{Exploded view of the camera assembly}, with eccentric rotating mass for induced vibration.}
\label{fig:mech_system}
\vspace{-2ex}
\end{figure}

%% file: figs/real_camera.tex
\begin{figure}[t!]
\centering
  \centering
  \includegraphics[width=\linewidth]{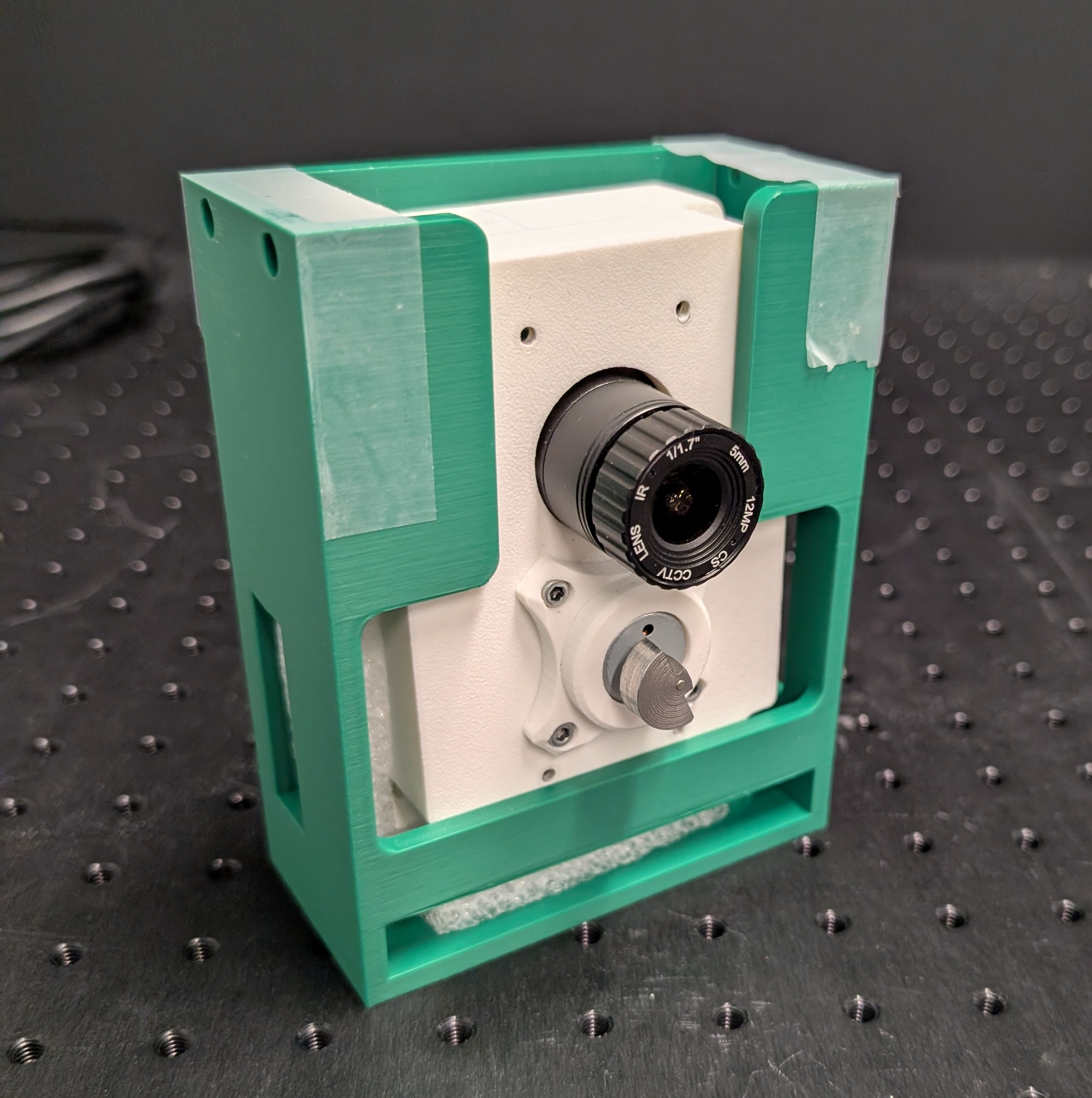}
\caption{\textbf{A photo of the hardware prototype.}}
\label{fig:real_camera}
\vspace{-2ex}
\end{figure}

%% file: figs/supplementary/data_collection_env.tex
\begin{figure}
\begin{subfigure}[T]{.33\linewidth}
    \centering
    \includegraphics[width=\linewidth]{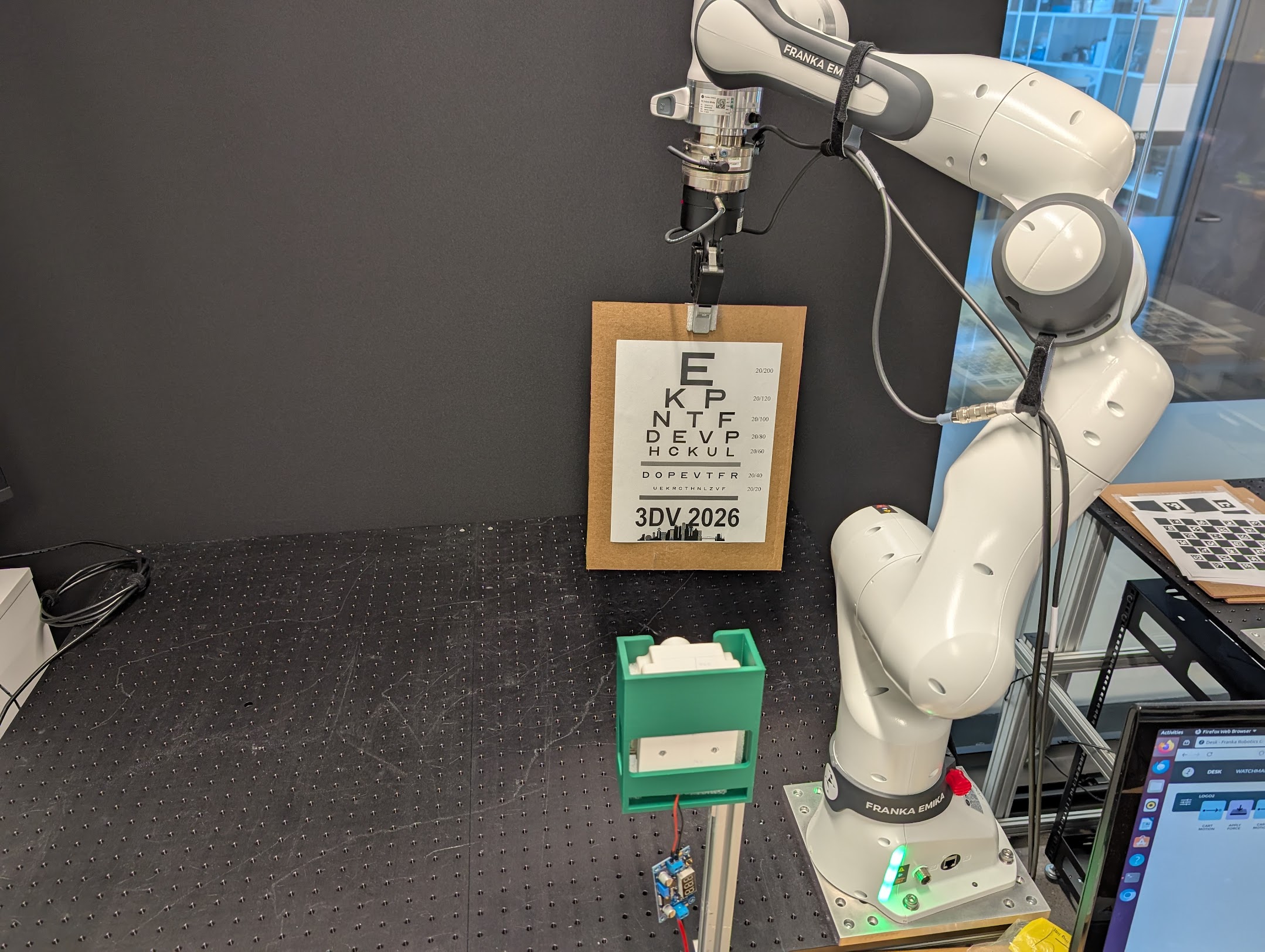}
    \end{subfigure}%
\begin{subfigure}[T]{.33\linewidth}
    \centering
    \includegraphics[width=\linewidth]{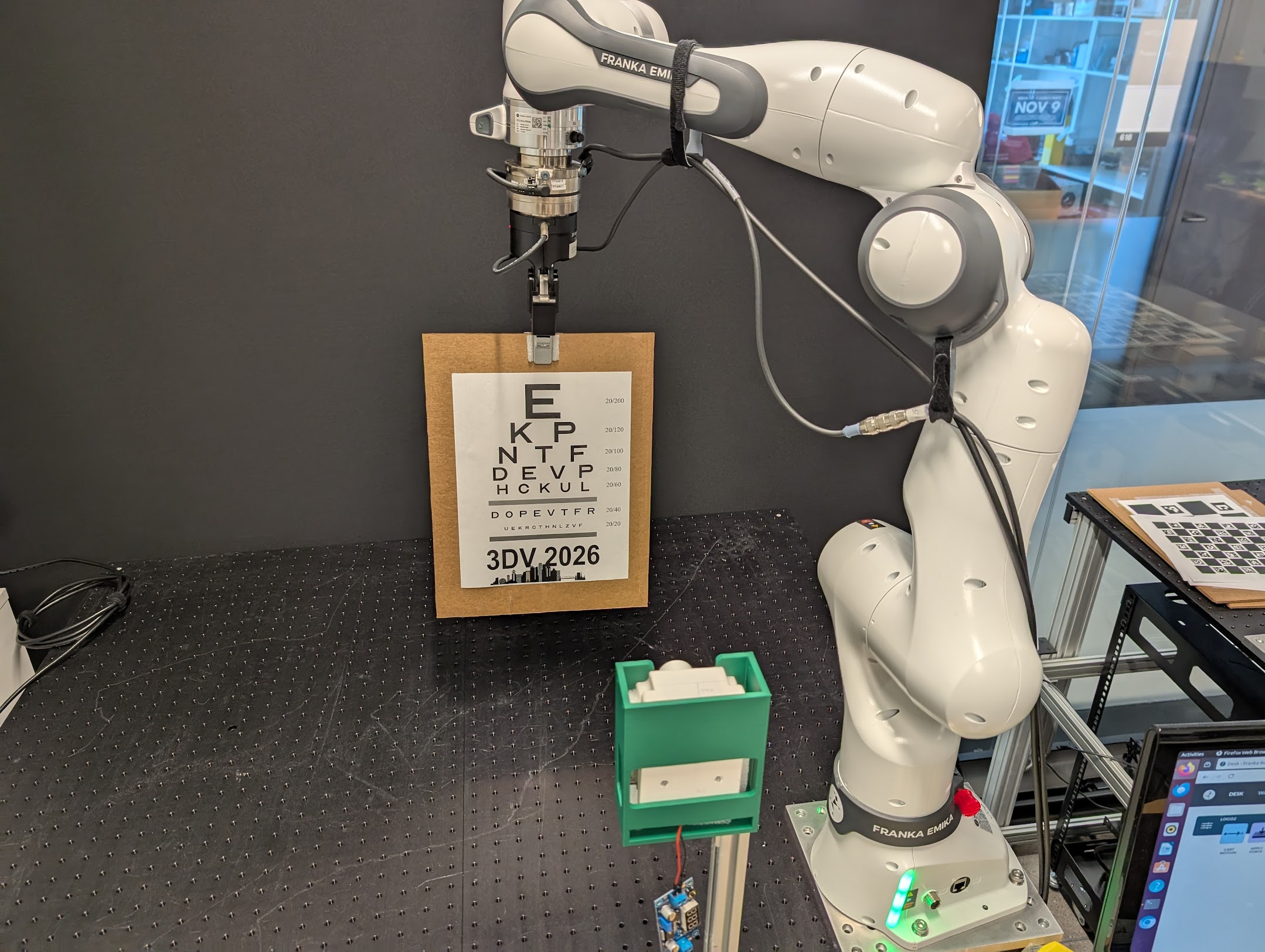}
    \end{subfigure}%
\begin{subfigure}[T]{.33\linewidth}
    \centering
    \includegraphics[width=\linewidth]{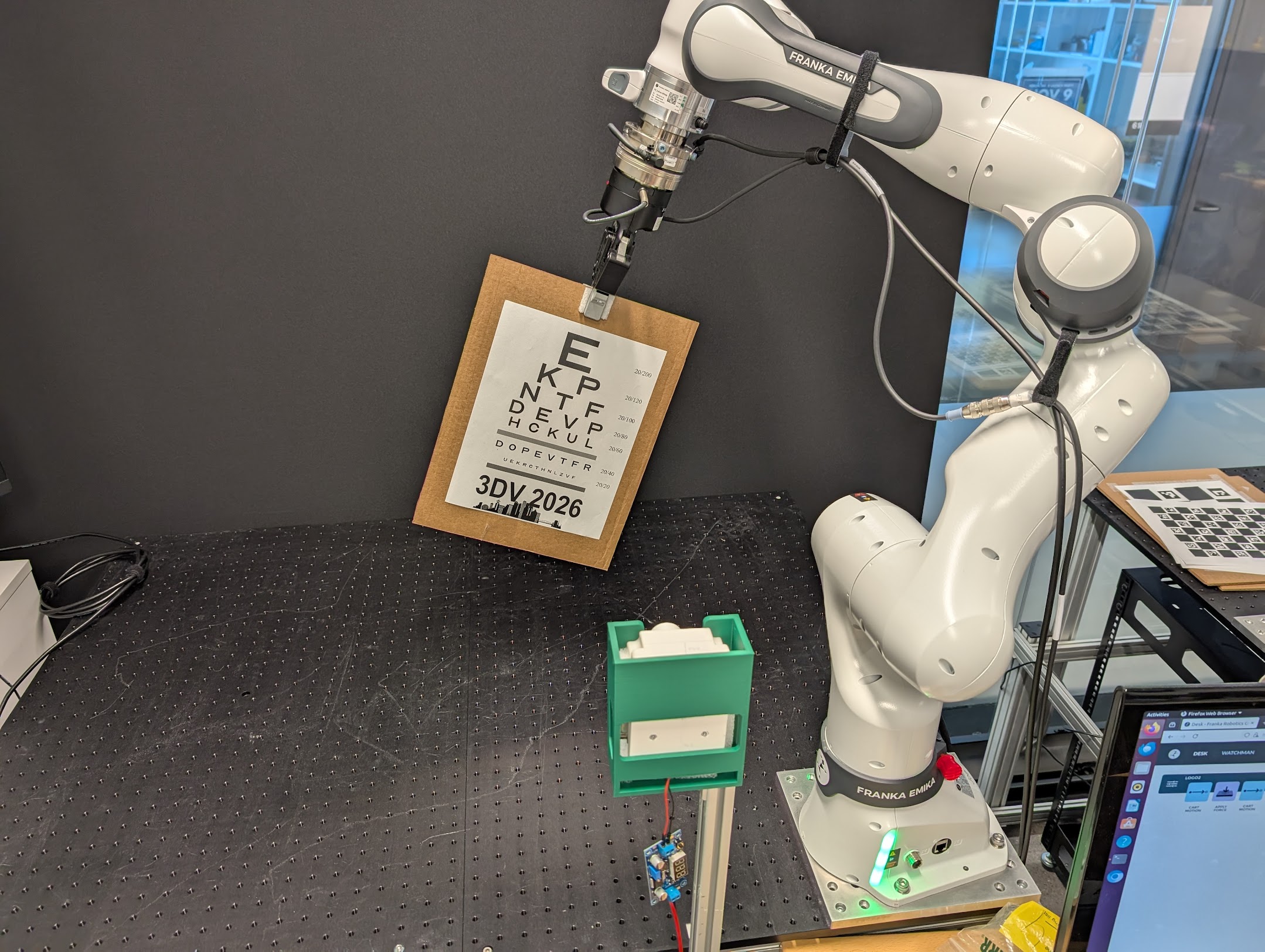}
    \end{subfigure}%
  \caption{\textbf{Data collection setup.} The Franka arm moves the board with the pattern in front of the prototype VIBES hardware.}
  \label{fig:data_collection_env}
\end{figure}

%% file: figs/supplementary/real_world_patterns.tex
\newbox{\bigpicturebox}
\begin{figure}
\sbox{\bigpicturebox}{%
  \begin{subfigure}[b]{.5\linewidth}
  \scalebox{1}[1.2]{\includegraphics[width=\linewidth]{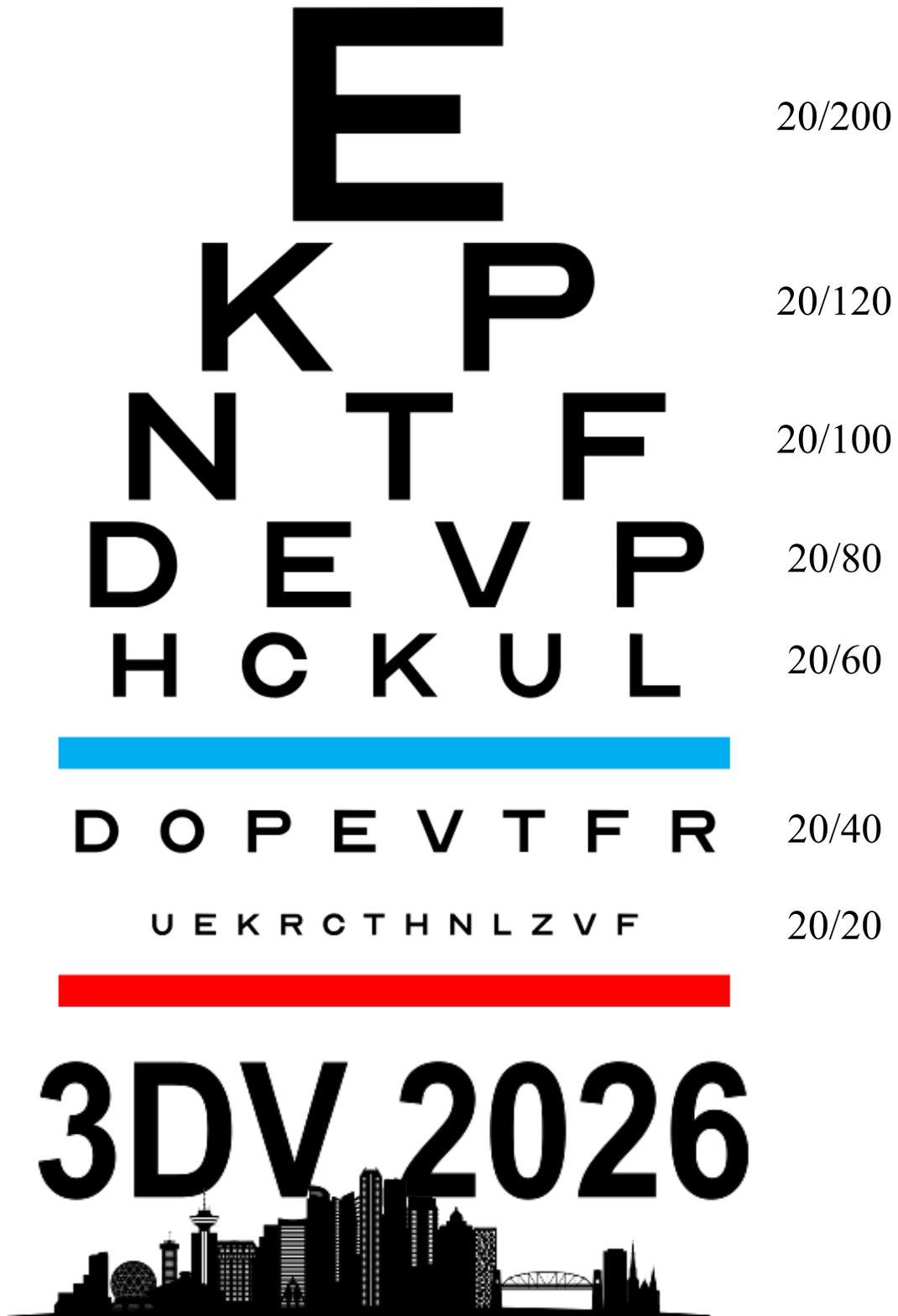}}
\caption{Logo}
\end{subfigure}%
}%
\usebox{\bigpicturebox}%
\begin{minipage}[b][\ht\bigpicturebox][s]{0.5\linewidth}
\begin{subfigure}{\linewidth}
\includegraphics[width=\linewidth]{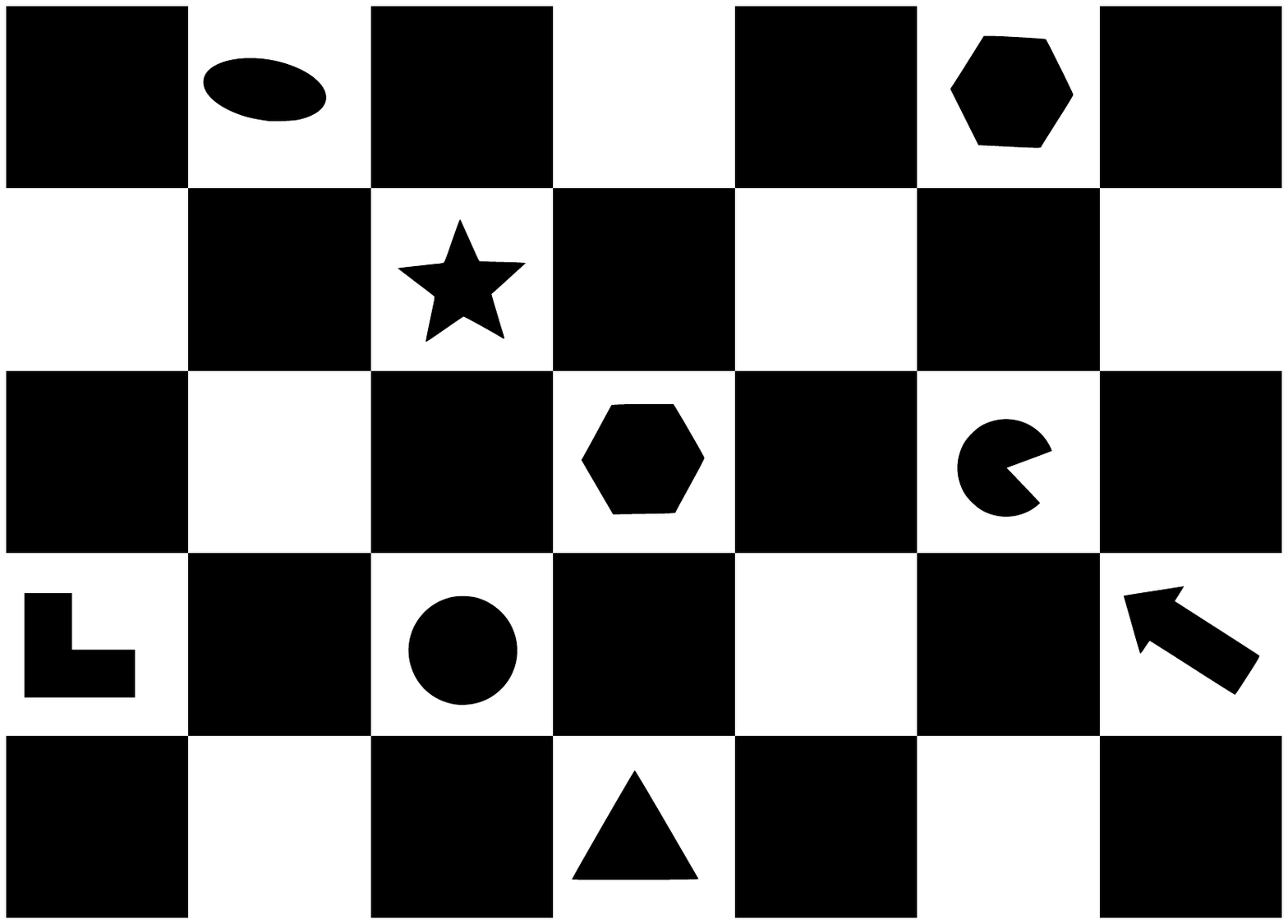}
\caption{Checkerpattern}
\end{subfigure}
\vfill
\begin{subfigure}[b]{\linewidth}
\includegraphics[width=\linewidth]{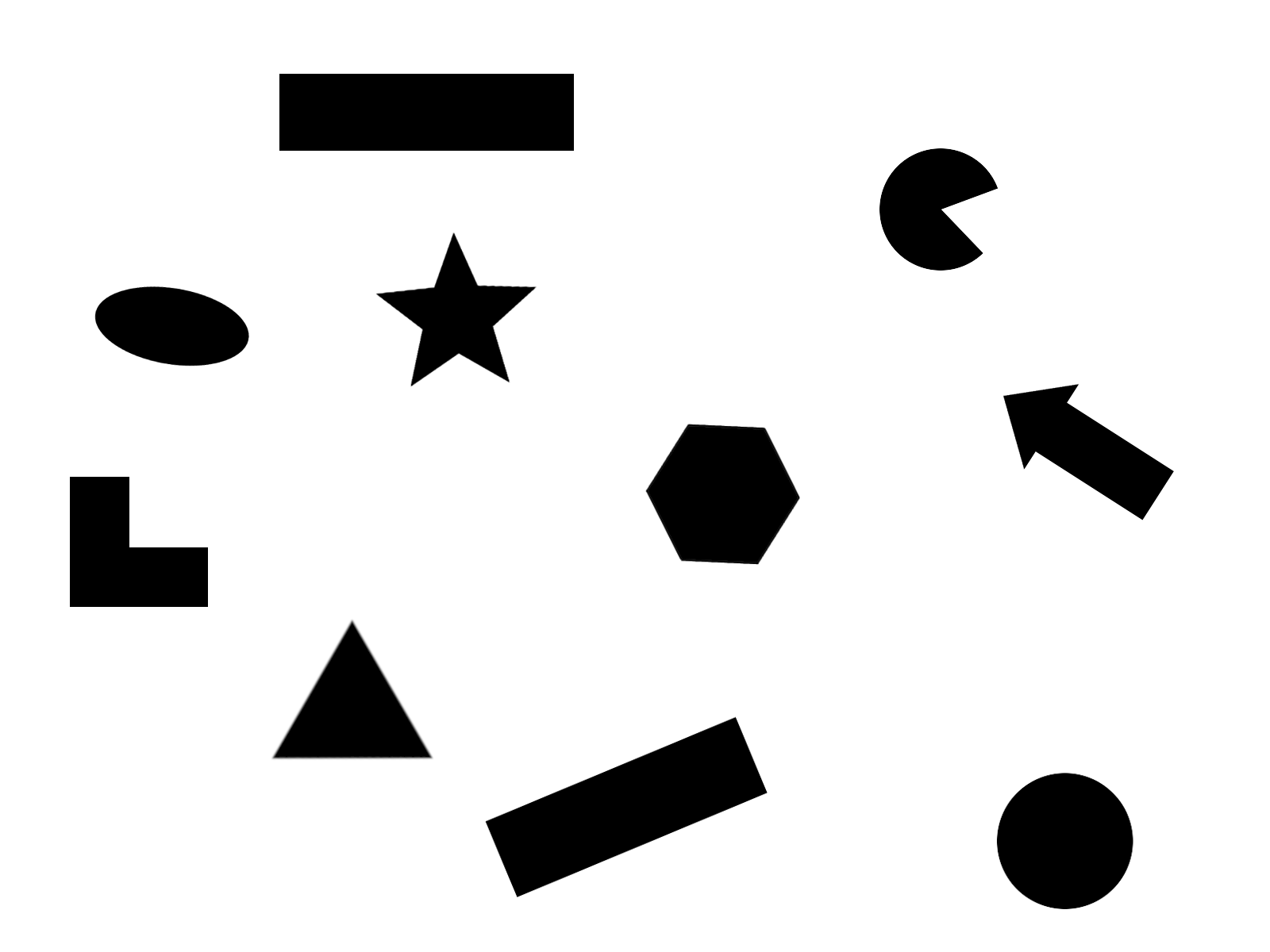}
\caption{Pattern}
\end{subfigure}
\end{minipage}
\caption{\textbf{The patterns used in the real-world dataset.}}
  \label{fig:real_world_patterns}
\end{figure}

%% file: tabs/supplementary/depth_stagger.tex
\begin{table}[t!]
\centering
\begin{tabular}{ccc}
\hline
Scene Ratio & Pattern 1 Distance & Pattern 2 Distance \\ \hline
0.33        & 21.0 cm            & 64.1 cm            \\
\colorrow 0.5         & 32.1 cm            & 64.1 cm            \\
0.66        & 40.6 cm            & 64.1 cm            \\ \hline
\end{tabular}
\caption{\textbf{Distances used in the synthetic depth estimation data.}}
\label{table:synthetic_depth_data}
\end{table}

%% file: tabs/supplementary/power_consumption.tex
\begin{table*}[t!]
\centering
\begin{tabular}{cccc}
\toprule
\textbf{Methodology} & \textbf{Motor Type} & \makecell{\textbf{Power} \\ \textbf{Consumption}} $\downarrow$  & \textbf{Measurement/Evaluation Basis} \\
\midrule
\methodname & DC Hobby Motor~\cite{hobby_motor_sparkfun} & \textbf{0.282W} & Full Power \\
\midrule
Orchard et al.~\cite{Orchard15fns} & Dynamixel MX-28T~\cite{dynamixel_mx_28t} & \colortab 2.4W & \colortab Standby \\
\midrule
\multirow{2}{*}{Testa et al.~\cite{testa2020dynamic} } & \multirow{2}{*}{PTU-D46-17~\cite{flir_ptu_d46}} & 6W & Low Power \\
& & \colortab 13W & \colortab  Full Power \\ 
\midrule
\multirow{2}{*}{Testa et al.~\cite{testa2023active}} & \multirow{2}{*}{PTU-E46~\cite{flir_ptu_e46}} & 6W &  Low Power \\
& & \colortab 13W & \colortab  Full Power \\
\midrule
AMI-EV~\cite{botao2024microsaccade} & DJI M2006 BLDC~\cite{dji_m2006_bldc} & 14.4W & No-load \\
\bottomrule
\end{tabular}
\caption{\textbf{Power Consumption}. \methodname facilitates persistent event generation, while using significantly less power than other methods.}
\vspace{-3ex}
\label{table:power_consumption}
\end{table*}

%% file: figs/supplementary/depth_amplitude.tex
\begin{figure}[!t]
    \centering
    \includegraphics[width=\linewidth]{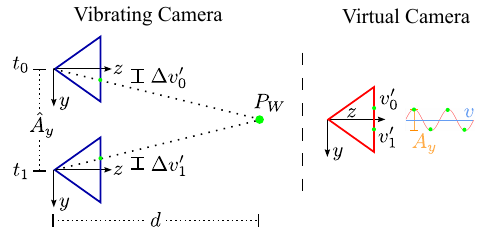} \hfill
    \vspace{-5ex}
    \caption{\textbf{Schematic representation of the induced stereo system.}  
The induced motion produces an effect analogous to a stereo system. The blue cameras are the vibrating camera at two different times, $t_0$ and $t_1$. The point $P_W$ is projected onto the camera plane, and at the two different timestamps has a different position $v'_0$ and $v'_1$ respectively. On the virtual camera, we want to recover the static omponent of the vibratonal motion indicated as $v$. While the baseline $\hat{A_y}$ is unknown, the amplitude of the sinusoidal motion $A_y$ can be estimated from the incoming event stream. This amplitude provides a strong cue for relative depth: if multiple tracks are initialized in the scene, the ratio between their amplitudes corresponds to their relative depth, as described in~\autoref{eq:rel_depth}.}
\label{fig:rel_depth_ampl}
\vspace{-2ex}
\end{figure}